%% file: banpick-tog.tex
\documentclass[journal]{IEEEtran}
\usepackage{times}  
\usepackage{helvet} 
\usepackage{courier}  
\usepackage[hyphens]{url}  
\usepackage{graphicx} 
\usepackage{caption} 
\usepackage{subfigure}
\usepackage{amsfonts,amsmath,amssymb,amsthm,mathtools}
\usepackage[T1]{fontenc}
\usepackage{color}
\usepackage{booktabs}
\definecolor{light-gray}{gray}{0.6}

\ifCLASSINFOpdf
\else
\fi
\begin{document}
%
\title{Which Heroes to Pick? Learning to Draft in MOBA Games with Neural Networks and Tree Search}
%
%
%

\author{Sheng Chen $^{*}$, 
        Menghui Zhu $^{*}$,
        Deheng Ye $^{\dag}$,
        Weinan Zhang,
        Qiang Fu,
        Wei Yang
\thanks{$^{*}$: equal contributions. \,\, $^{\dag}$: corresponding author.}
\thanks{Sheng Chen, Deheng Ye, Qiang Fu and Wei Yang are with the Tencent AI Lab, Shenzhen, China. (e-mail: victchen@tencent.com, dericye@tencent.com, leonfu@tencent.com, willyang@tencent.com)}
\thanks{Menghui Zhu and Weinan Zhang are with the Department
of Computer Science and Engineering, Shanghai Jiao Tong University, Shanghai, China. (e-mail: zerozmi7@sjtu.edu.cn, wnzhang@sjtu.edu.cn) Menghui Zhu was an intern in Tencent when this work was done.}
}

\input{IEEEtran/commands}
\maketitle

\begin{abstract}
Hero drafting is essential in MOBA game playing as it builds the team of each side and directly affects the match outcome. 
State-of-the-art drafting methods fail to consider: 1) drafting efficiency when the hero pool is expanded;
2) the multi-round nature of a MOBA 5v5 match series, i.e., two teams play best-of-$N$ and the same hero is only allowed to be drafted once throughout the series. 
In this paper, we formulate the drafting process as a multi-round combinatorial game and propose a novel drafting algorithm based on neural networks and Monte-Carlo tree search, named JueWuDraft.
Specifically, we design a long-term value estimation mechanism to handle the best-of-$N$ drafting case. 
Taking \textit{Honor of Kings}, one of the most popular MOBA games at present, as a running case, we demonstrate the practicality and effectiveness of JueWuDraft when compared to state-of-the-art drafting methods. 
\end{abstract}


%
\IEEEpeerreviewmaketitle

\input{IEEEtran/paper/introduction}
\input{IEEEtran/paper/relatedwork}
\input{IEEEtran/paper/methodology}
\input{IEEEtran/paper/experiment}

\input{IEEEtran/paper/conclusion}


%



\section*{Acknowledgment}
This work is supported by the ``New Generation of AI 2030'' Major Project (2018AAA0100900), Shanghai Municipal Science and Technology Major Project (2021SHZDZX0102), National Natural Science Foundation of China (62076161, 61632017) and the Tencent Rhino-Bird Research Project.


\ifCLASSOPTIONcaptionsoff
  \newpage
\fi



%
\bibliographystyle{IEEEtran}
\input{banpick-tog.bbl}




%

\begin{IEEEbiography}[{\includegraphics[width=1in,height=1.25in,clip,keepaspectratio]{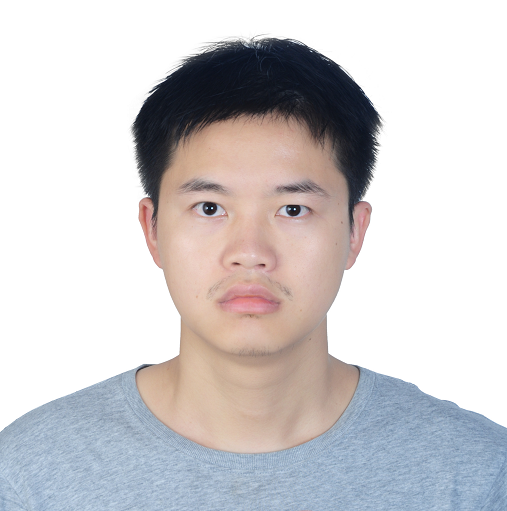}}]{Sheng Chen}
received the B.S. and M.S. degrees from the University of Science and Technology of China, Hefei, China, in 2015 and 2018, respectively. 

He is a Researcher with the Tencent AI Lab, Shenzhen, China. His research interests include Game AI and applied machine learning.
\end{IEEEbiography}
\begin{IEEEbiography}[{\includegraphics[width=1in,height=1.25in,clip,keepaspectratio]{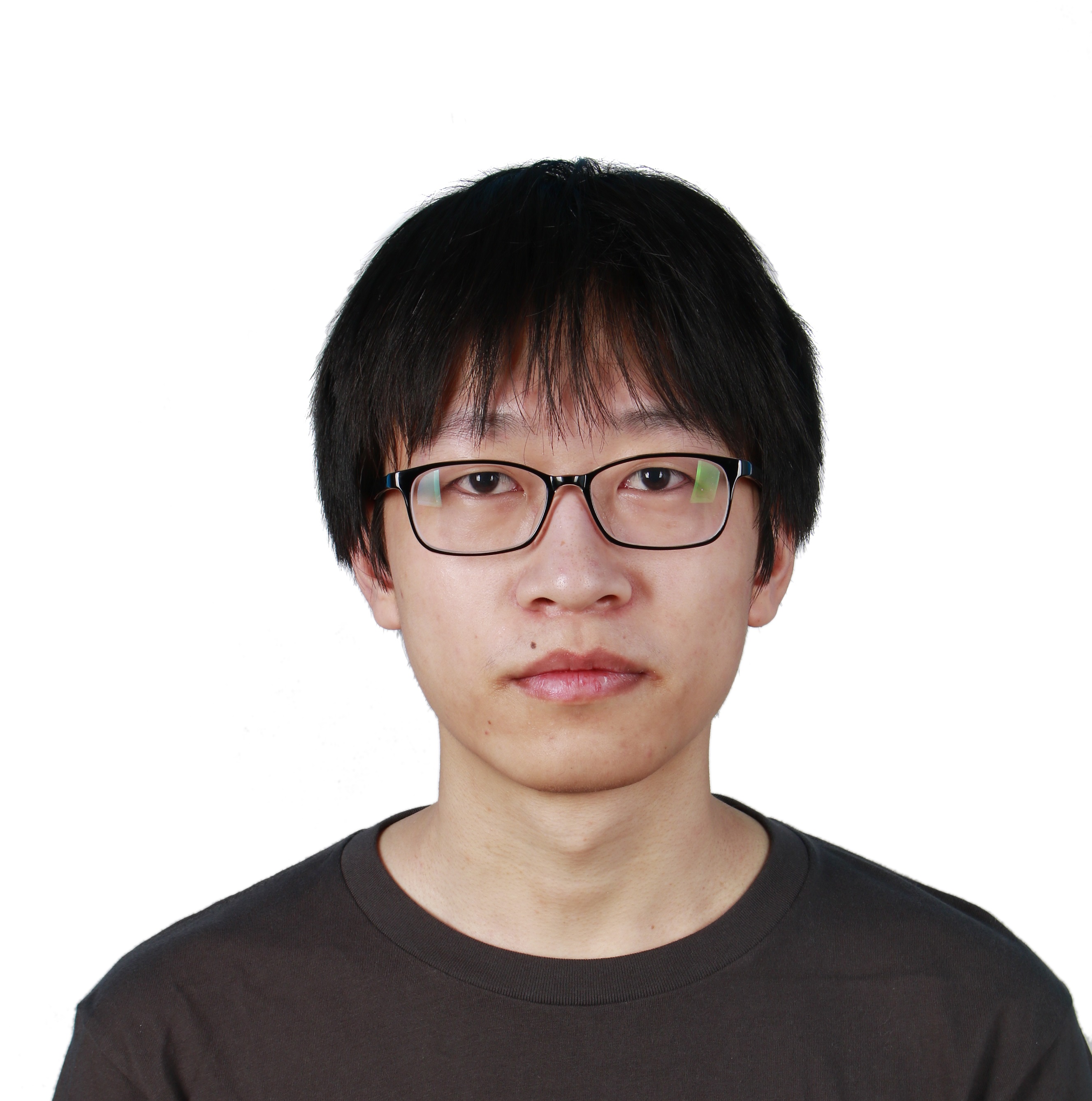}}]{Menghui Zhu}
received the B.S. from the Department of Computer Science and Engineering, Shanghai Jiao Tong University, in 2020.

He is a postgraduate student in Shanghai Jiao Tong University. His research interests lie primarily in  reinforcement learning and its applications.
\end{IEEEbiography}
\begin{IEEEbiography}[{\includegraphics[width=1in,height=1.25in,clip,keepaspectratio]{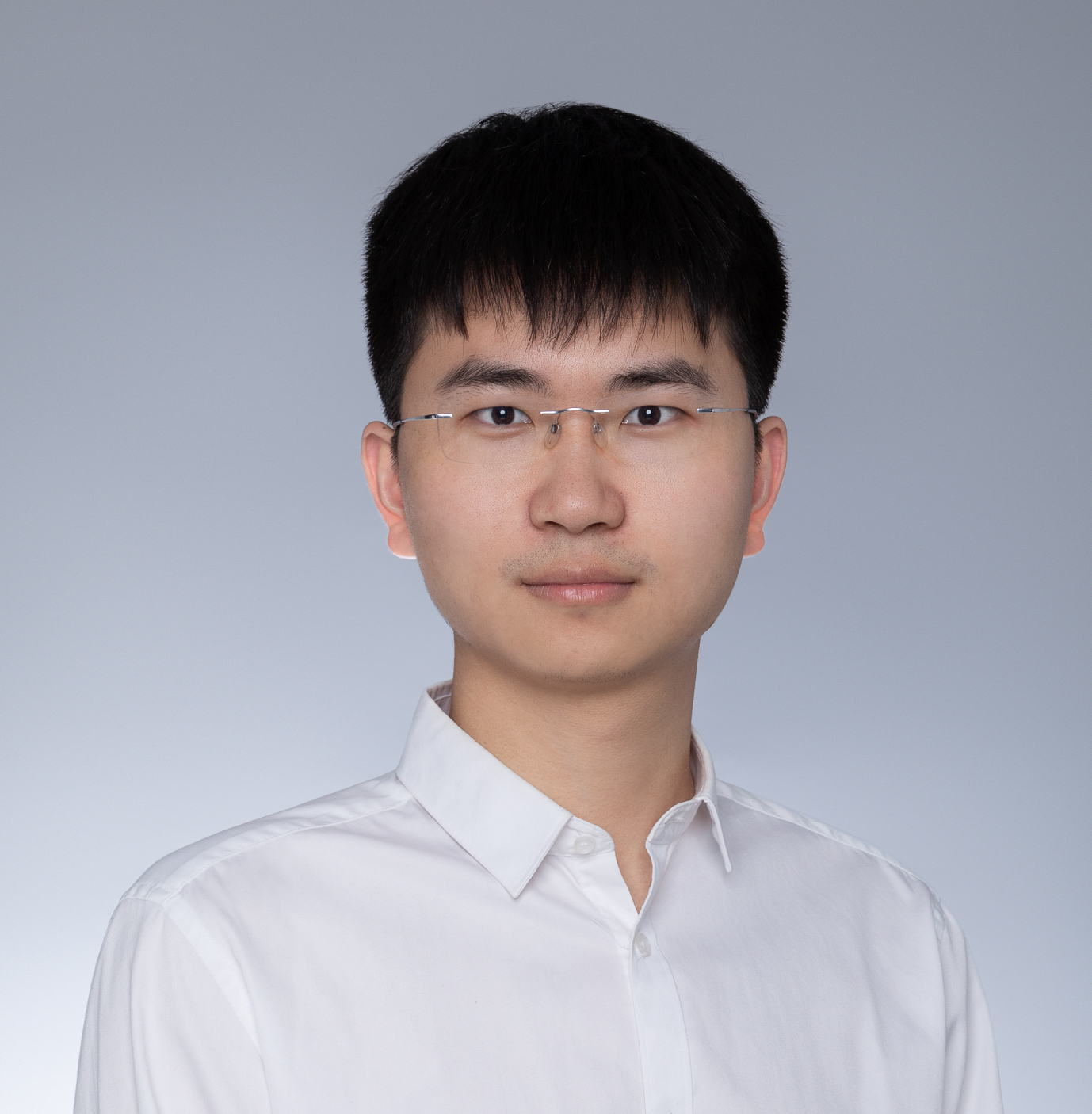}}] {Deheng Ye} finished his Ph.D. from the School of Computer Science and Engineering, Nanyang Technological University, Singapore, in 2016. 

He is now a Researcher and Team Manager with the Tencent AI Lab, Shenzhen, China, where he leads a group of engineers and researchers developing large-scale learning platforms and intelligent AI agents. 
He is broadly interested in applied machine learning, reinforcement learning, and software engineering. 
He has been serving as a (senior) PC for NeurIPS, ICLR, AAAI, IJCAI, etc. 
\end{IEEEbiography}
\begin{IEEEbiography}[{\includegraphics[width=1in,height=1.25in,clip,keepaspectratio]{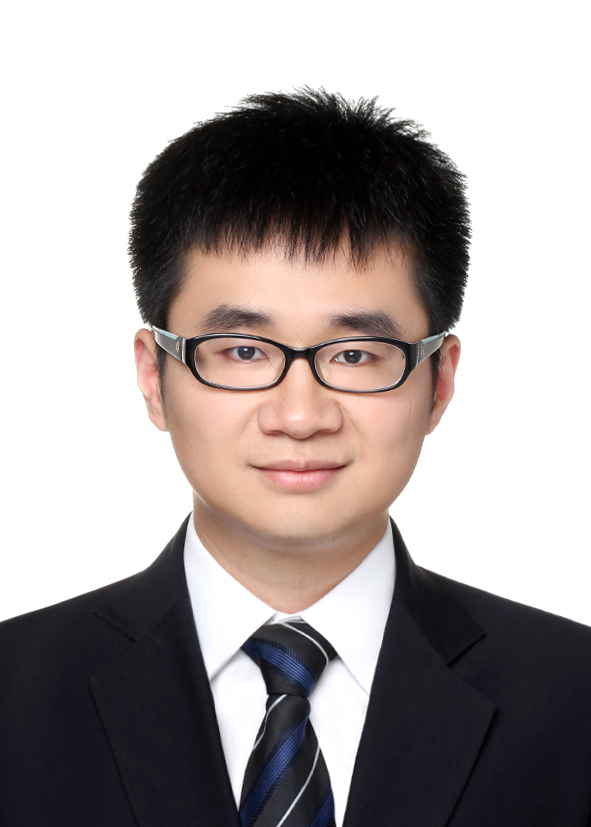}}]{Weinan Zhang}
is now an associate professor at Shanghai Jiao Tong University. He received his Ph.D. from the Computer Science Department of University College London in 2016. His research interests include reinforcement learning, deep learning and data science with various real-world applications of recommender systems, search engines, text mining and generation, game AI etc. He has published over 100 research papers on international conferences and journals and has been serving as a (senior) PC member at ICML, NeurIPS, ICLR, KDD, AAAI, IJCAI, SIGIR etc. and a reviewer at JMLR, TOIS, TKDE, TIST etc. His research work achieves Best Paper Honorable Mention Award in SIGIR 2017, ACM Shanghai Rising Star Award 2017 and Best System Paper Award at CoRL 2020.
\end{IEEEbiography}
\begin{IEEEbiography}[{\includegraphics[width=1in,height=1.25in,clip,keepaspectratio]{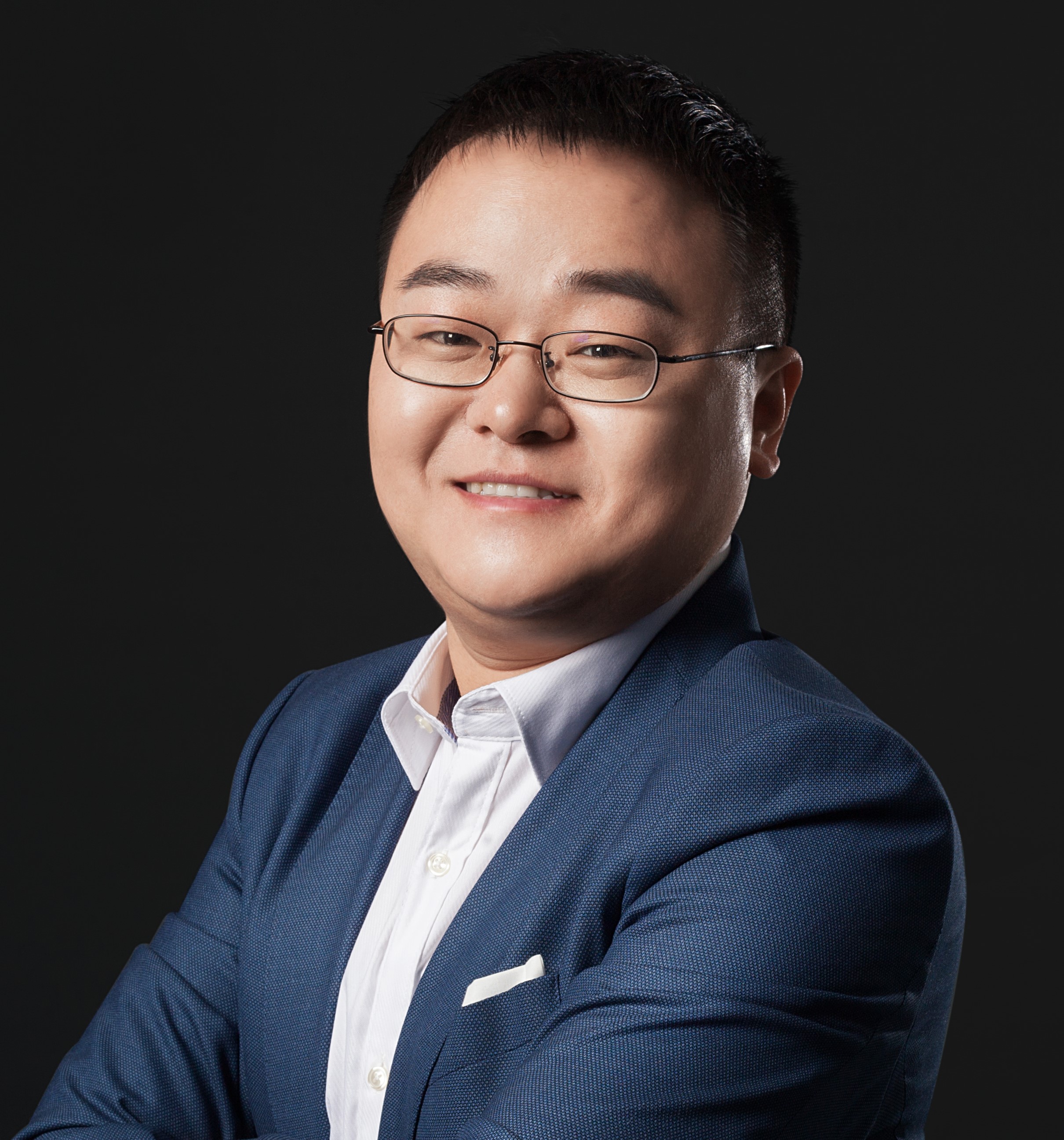}}]{Qiang Fu}
received the B.S. and M.S. degrees from the University of Science and Technology of China, Hefei, China, in 2006 and 2009, respectively. He is the Director of the Game AI Center, Tencent AI Lab, Shenzhen, China. 

He has been dedicated to machine learning, data mining, and information retrieval for over a decade. His current research focus is game intelligence and its applications, leveraging deep learning, domain data analysis, reinforcement learning, and game theory.
\end{IEEEbiography}
\begin{IEEEbiography}[{\includegraphics[width=1in,height=1.25in,clip,keepaspectratio]{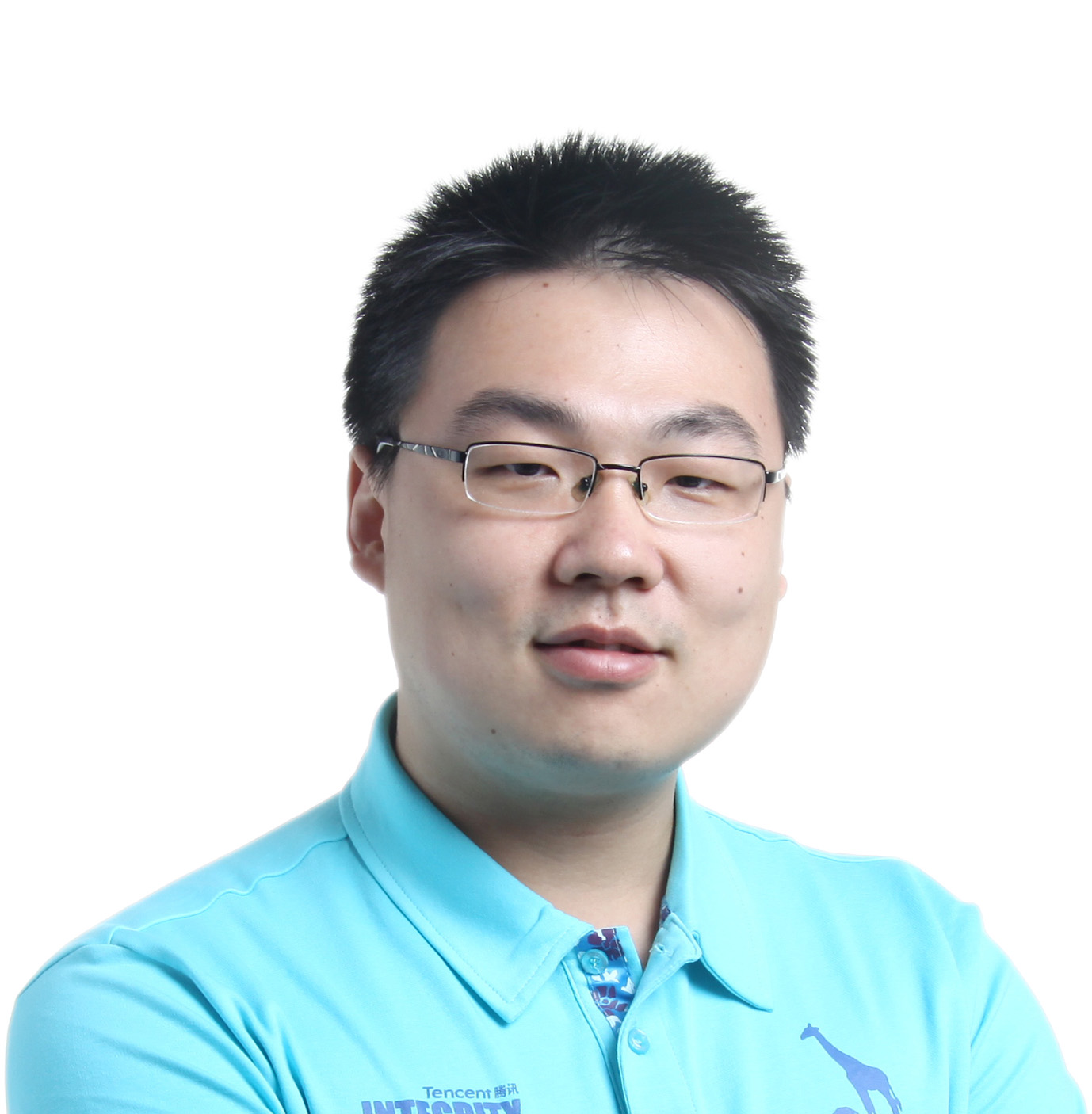}}]{Wei Yang}
received the M.S. degree from the Huazhong University of Science and Technology, Wuhan, China, in 2007. He is currently the General Manager of the Tencent AI Lab, Shenzhen, China. 

He has pioneered many influential projects in Tencent in a wide range of domains, covering Game AI, Medical AI, search, data mining, large-scale learning systems, and so on.
\end{IEEEbiography}





\end{document}

%% file: IEEEtran/commands.tex
\newcommand{\fig}[1]{Fig.~\ref{#1}}
\newcommand{\eq}[1]{Eq.~(\ref{#1})}
\newcommand{\tb}[1]{Tab.~\ref{#1}}
\newcommand{\se}[1]{Section~\ref{#1}}
\newcommand{\ap}[1]{Appendix~\ref{#1}}
\newcommand{\pa}[1]{Part~\ref{#1}}
\newcommand{\lm}[1]{Lemma~\ref{#1}}
\newcommand{\prop}[1]{Proposition~\ref{#1}}
\newcommand{\alg}[1]{Algo.~\ref{#1}}
\newcommand{\theo}[1]{Theorem~\ref{#1}}
\newcommand{\defi}[1]{Definition~\ref{#1}}
\newcommand{\minisection}[1]{\vspace{3pt}\noindent\textbf{#1.}}

%% file: IEEEtran/paper/introduction.tex
\section{Introduction} \label{sec:intro}
Artificial Intelligence for games (Game AI) has drawn a lot of attention since the AlphaGo and AlphaZero defeated human professionals in board games \cite{silver2016mastering, silver2017mastering}. Also, we have witnessed the success of AI agents in other game types, including Atari series \cite{mnih2015human}, first-person shooting (FPS) games like \textit{Doom} \cite{lample2017playing}, video games like \textit{Super Smash Bros} \cite{chen2017game}, card games like \textit{Poker} \cite{brown2018superhuman}, etc. Despite their great success, 
another general type of game -- real-time strategy (RTS) video games \cite{buro2003real} still has not been conquered by AI agents, due to its raw complexity
and multi-agent challenges. Thus, recently researchers have paid more attention to RTS games such as \textit{Defense of the Ancients 2 (Dota2)} \cite{berner2019dota, brockman2018openai}, \textit{StarCraft \uppercase\expandafter{\romannumeral2}} \cite{vinyals2017starcraft,tian2017elf,sun2018tstarbots,vinyals2019grandmaster} and \textit{Honor of Kings} \cite{ye2020supervised,ye2019mastering,ye2020towards}. 

As a subgenre of RTS games, 
Multiplayer Online Battle Arena (MOBA) is one of the most popular contemporary
e-sports \cite{berner2019dota,ye2020supervised,ye2019mastering,ye2020towards}. Due to its playing mechanics,  which involve multi-agent competition and cooperation, imperfect information, complex action control, enormous state-action space, MOBA is considered as a interesting testbed for AI research \cite{silva2017moba,robertson2014review}.
The standard game mode in MOBA is 5v5, i.e., a team of five players competes against another team of five players. 
The ultimate objective is to destroy the home base of the opposite team. 
Each player controls a single game character, known as hero, to cooperate with other teammates in attacking opponents' heroes, minions, turrets, and neutral creeps, while defending their own in-game properties. 

\begin{figure}[htbp]
    \centering
    \includegraphics[width=0.48\textwidth,height=0.25\textheight]{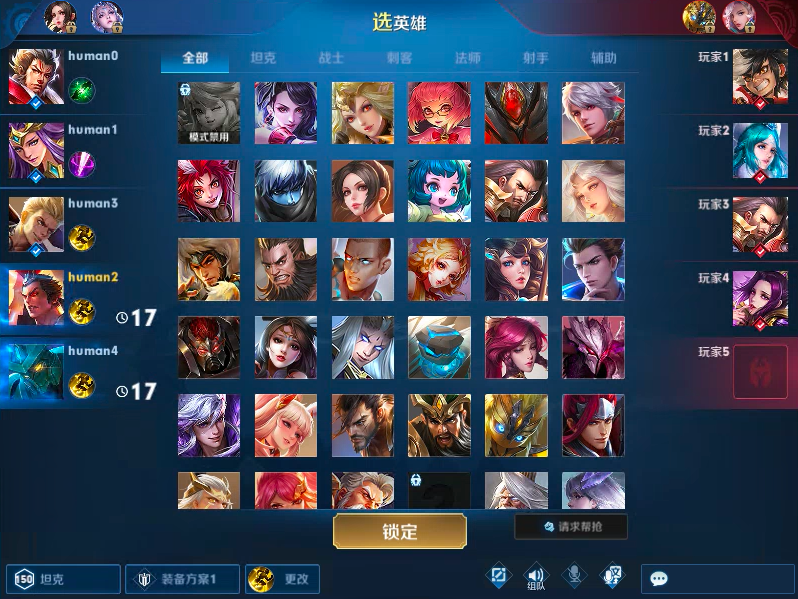}
    \caption{Screenshot of hero drafting phase in \textit{Honor of Kings}. The left column shows heroes picked by ally camp, and the right column shows heroes picked by enemy camp. Drafting alternates between two teams until all players have selected one heroes, which makes up a complete lineup.}
    \label{fig:sreenshoot}
\end{figure}

A single MOBA match consists of two phases: (i) the pre-match phase during which the 10 players of the two teams select heroes from a pool of heroes, a.k.a., hero drafting or hero picking, and (ii) the in-match phase where two teams start to fight until the game ends. Here we provide a screenshot of hero drafting phase in \textit{Honor of Kings} in \fig{fig:sreenshoot}.
Drafting alternates between two teams until each player has selected one hero. The pick order is “1-2-2-1-1-2-2-1-1-2”, meaning that the first team picks one hero, followed by the second team picking two heroes, then the first team picking two heroes, and so on. More details will be explained in following chapters.
We refer to the 10 heroes in a completed draft as a lineup, which directly affects future strategies and match results. Thus, hero drafting process is very important and a necessary part of building a complete AI system to play MOBA games \cite{berner2019dota,ye2020towards}.

To determine the winner of two teams, the best-of-$N$ rounds, where the odd number $N$ can be 1, 3, 5, etc., is the widely used rule, which means maximum $N$ rounds can be played until one team wins $(N+1)/2$ rounds. 

Each MOBA hero in one team is uniquely designed to have hero-specific skills and abilities, which together add to a team's overall power. 
Also, heroes have the sophisticated counter and supplement relationships to each other. For example, in \textit{Dota 2}, hero \textit{Anti-Mage}'s skill can reduce an opponent hero's mana resource, making him a natural counter to hero \textit{Medusa}, whose durability heavily relies on the amount of mana. 
For another example, in \textit{Honor of Kings}, hero \textit{MingShiyin} can enhance the normal attack of ally heroes, making him a strong supplement to the shooter hero \textit{HouYi}. 
Therefore, to win the match, players need to carefully draft heroes that can enhance the strengths and compensate for the weaknesses of teammates' heroes, with respect to the opposite heroes. 

In MOBA games like \textit{Honor of Kings} and \textit{Dota 2}, there are possibly more than 100 heroes that can be selected by a player during drafting. For example, the latest version of \textit{Honor of Kings} includes 102 heroes. Thus the number of possible hero lineups can be approximately $5.37 \times 10^{15}$ ($C_{102}^{10} \times C_{10}^{5}$, i.e., picking 10 heroes out of a pool of 102 heroes and picking 5 out of 10).
Due to the complex relationships among heroes as mentioned above and the huge number of possible lineups, drafting a suitable hero that can synergize teammates and counter opponents is challenging to human players. 

State-of-the-art methods for MOBA hero drafting are from OpenAI Five \cite{berner2019dota} and DraftArtist \cite{chen2018art}. 
OpenAI Five, the AI program for playing \textit{Dota 2}, uses the Minimax algorithm \cite{fan1953minimax,sion1958general} for hero drafting since only 17 heroes were supported. 
However, Minimax constructs a complete search tree comprising of all possible picks alternated by each player, making it computationally intractable to scale to large hero pools, e.g., 100 heroes. With pruning techniques \cite{knuth1975analysis}, the computation time of Minimax algorithm is reduced by preventing exploring useless branches of the search tree. However, its efficiency improvement is not sufficient when the hero pool is large.
DraftArtist leverages Monte-Carlo Tree Search (MCTS) \cite{coulom2006efficient, kocsis2006bandit} to estimate the value of each pick by simulating possible following picks until the lineup completes. However, it only considers one single match only, i.e., best-of-1, which may result in less optimal drafting result. Moreover, its MCTS simulation adopts random rollouts to get the reward, which may need a large number of iterations to get an accurate estimation. Thus, it's inefficient. 

In this paper, we characterize the MOBA drafting process as a two-player zero-sum game with perfect information. The aim of each player is to maximize the winning rate of its team consisting of five heroes drafted from a large pool of heroes against the opponent five-hero team, under the condition that one hero is only allowed to be picked once for the same team. 
To handle with this problem, we propose to combine Monte-Carlo Tree Search (MCTS) and Neural Networks (NN). 
Specifically, we use MCTS with a policy and value network, where the value network is to predict the value of the current state directly instead of the inefficient random roll-out to get the reward. 
Unlike board games such as \textit{Chess} and \textit{Go} in which the terminal state determines the winner of the match, the end of the drafting process is not the end of a MOBA match. Therefore, we cannot get match results directly. 
To deal with this, we build a neural predictor to predict the winning rate of particular lineups, named as win-rate predictor. The win-rate predictor is trained in the way of supervised learning with a match dataset. The winning rate of the terminal state is used as the reward for back-propagation of MCTS and training the value network.

In best-of-$N$ rule, there are several rounds to determine the final winner.
Drafting a hero for the current round will affect the outcome of later rounds. Specifically, players are not allowed to pick heroes that were picked by themselves in previous rounds. 
To handle this problem, we formulate the drafting process as a multi-round combinatorial game and design a long-term value mechanism. Thus, our drafting strategy will have a longer consideration for both current rounds and following rounds.  

To sum up, we make the following contributions: 
\begin{itemize}
    \item We propose a hero drafting method called JueWuDraft for MOBA games via leveraging neural networks and Monte-Carlo tree search. Specifically, we use MCTS with a policy and value network, where the value network is to evaluate the value of the current state, and the policy network is used for sampling actions for next drafting.
    \item We formulate the best-of-$N$ drafting problem as a multi-round combinatorial game, where each round has a winning rate for its final lineup. To adapt to such best-of-$N$ drafting problem, we design a long-term value mechanism. Thus the value estimation of the current state will consider longer for the following rounds.
    \item We conduct extensive experiments and the results demonstrate the superiority of JueWuDraft over state-of-the-art drafting methods on both single-round and multi-round games. Specifically, JueWuDraft performs better on multi-round games, which indicates our method can handle further consideration better than other strategies.
\end{itemize}


%% file: IEEEtran/paper/relatedwork.tex
\section{Related Work}

Previous methods for MOBA hero drafting can be categorized into the following four types. 

1) Historical selection frequency.
Along this line, Summerville et al. \cite{summerville2016draft} proposed to recommend heroes that players tend to select. Specifically, hero drafting is modeled as a sequence prediction problem, in which the next hero to pick is predicted from the sequence of picks made already in the draft. Since the sequence prediction model is trained with supervised learning from pick sequences in historical matches and does not consider their win-loss results, the predicted hero is ``what is likely to be picked, not what is necessarily best'' \cite{summerville2016draft}.  
In other words, this method will pick heroes with the highest frequencies rather than heroes with the highest winning rates, which may not be optimal for the team victory.

2) Previous winning rate based. In \cite{hanke2017recommender}, Hanke and Chaimowicz proposed to mine association rules \cite{agarwal1994fast} from historical hero lineup data and use it as heuristics to recommend the next hero to pick. Here the association rules are hero subsets that appear frequently together either in the winning team or in opposite teams. Then heroes mined by association rules are considered as a recommendation list for the next drafting step. However, such a method does not consider which heroes to be picked in the rest of the drafting process, thus 
it is essentially a myopic, greedy-based approach.

3) Minimax algorithm. OpenAI Five uses the Minimax algorithm for hero drafting \cite{berner2019dota}, with 17 heroes available (only 4,900,896 combinations of two 5-hero teams from the pool of 17 heroes). Minimax method will make the move that maximizes the minimum value of the position resulting from the opponent's possible following moves. Specifically, a hero is picked that can maximize the worst case scenario after the opponent hero selection. However, the time complexity of this method increases at the polynomial scale when the hero pool enlarges. Even with alpha-beta pruning technique \cite{knuth1975analysis}, the improvement of time complexity is limited. Given a large hero pool, e.g., 100 heroes (more than $10^{15}$ combinations), the method used for drafting in OpenAI Five will be computationally intractable.

4) Monte-Carlo tree search (MCTS). MCTS \cite{coulom2006efficient} can be seen as an alternative to Minimax algorithm, which improves the efficiency of search by using random roll-outs for value approximations. Recently, it has attracted much attention because of its successful application to Go-playing programs \cite{silver2016mastering,silver2017mastering}, General Game Playing \cite{finnsson2008simulation}, and Real-Time Strategy games \cite{balla2009uct}.
Specifically, DraftArtist \cite{chen2018art} uses MCTS for estimating the values of hero combinations, which is the state-of-the-art hero drafting algorithm.  However, this method fails to consider the fact that a formal MOBA match contains a series of $N$ rounds. Besides, random rollouts adopted by MCTS simulation are not efficient enough.

Although there are other works using machine learning models to predict match outcomes \cite{semenov2016performance, yang2014identifying}, they do not focus on how to utilize these models for hero drafting.

The combination of the planning ability from MCTS and the generalization ability from neural networks brings the agent to the level of beating human professionals in the game of \textit{Go} \cite{silver2016mastering}, as known as AlphaGo. Subsequently, a more general deep reinforcement learning (DRL) method is further applied to the \textit{Chess} and \textit{Shogi} 1v1 games \cite{silver2017mastering}, as known as AlphaZero. Different with AlphaGo, AlphaZero starts from zero with self-play without human demonstrations.

Similar to the board games like \textit{Go}, the drafting problem can also be treated as a two-player zero-sum perfect information game. Thus, we propose our drafting method based on MCTS and neural networks, which is similar to AlphaZero. Unlike board games, the end of drafting process can only determine the final lineup rather than win-lose information, thus we build a winning rate predictor to predict the winning rate of
particular lineups, which is used as the reward for back-propagation of MCTS and training the value network. Moreover, we formulate the best-of-$N$ drafting process as a multi-round combinatorial game. And we design a long-term value mechanism for value credit assignment \cite{sutton1984temporal}, which is normal in multi-agent reinforcement learning (MARL) field \cite{sunehag2017value,foerster2018counterfactual}, and also has been considered in MCTS/UCT \cite{kocsis2006bandit,vodopivec2017monte}. With such value design, our drafting method can take a long-term consideration for both the current round and the following rounds.

%% file: IEEEtran/paper/methodology.tex
\section{Method}

In this section, we describe JueWuDraft in detail, including (i) the problem formulation of the hero drafting process as a multi-round combinatorial game, (ii) how we apply MCTS and neural network for the optimal hero drafting, and (iii) how to design the long-term value mechanism.

\subsection{Problem Formulation}
As described above, drafting heroes plays an important role in MOBA, especially in complicated game matches. 
In formal MOBA matches, the best-of-$N$ round format is widely used to determine the winner, where $N$ is an odd, usually BO3 or BO5.  
For example, The competition format of \textit{King Pro League} (KPL) \footnote{KPL is the professional player league of Honor of Kings, which organizes annual esports tournament. Other popular MOBA games also have similar annual esports tournament, e.g., TI (The International) of Dota, LPL (LOL Pro League) of League of Legends.} always contains several rounds. 
Under this format, two teams fight against each other until one wins $(N+1)/2$ rounds (2 in BO3, 3 in BO5). Besides, players are not allowed to pick heroes that were picked by themselves in previous rounds of the multi-round games. 
It means that to win a multi-round game, one must take the state of both the current round and the following rounds into account during drafting. 

Therefore, instead of considering the drafting in each round individually, we define the draft in a full game $\mathcal{G}$ as \textit{a two-player zero-sum perfect information game} with repeated similar sub-structure in this paper. The game we defined consists of the following elements (in this paper we do not consider banning heroes):

\begin{itemize}
    \item The number of players in a full game: $P=2$, including $\{player1, player2\}$.
    
    \item The number of camps in a game: $M=2$, including $\{camp1, camp2\}$. Noted that either player can be $camp1$ or $camp2$, and $camp1$ always chooses heroes firstly.
    
    \item The number of rounds in a full game: $D$. It represents that a full game contains $D$ rounds, which means to win a full game, a player must win at least $\lceil D/2 \rceil$ rounds.
    
    \item The length of a round $L$: It defines that a round in a full game will last $L$ time-steps, there is one hero drafted at each time-step.
    
    \item Player order $\mathcal{O}_{round}$ and $\mathcal{O}_{first}$: For example, in \textit{Honor of Kings}, in every round, the two players take actions in the pick order $\mathcal{O}_{round}=\{1,2,2,1,1,2,2,1,1,2\}_{l}$. It represents that $camp1$ first picks a hero, and $camp2$ consecutively picks two heroes, then $camp1$ goes on to pick 2 heroes and so on. Another player order $\mathcal{O}_{first} = \{player_0, player_1, \ldots \}_{d}$ defines which player to pick first in the corresponding round. In the rule of KPL, the $\mathcal{O}_{first}$ is fixed to $\mathcal{O}_{first}=\{player1, player2, player1, \ldots \}_{d}$, where the two players take turns to pick firstly in the round.
    
    \item The turn function $\rho: t \rightarrow \{player1, player2\}$ : $\rho$ decides which player's turn to give an action according to the time-step $t$. Obviously, it is based on $\mathcal{O}_{round}$ and $\mathcal{O}_{first}$ as stated above.
    
    \item The state of the game $s \in \mathcal{S} \subset [0, N_{classes}]^{L \times D}$. The state $s$ is a vector of positive integers, of which the length is $L \times D$. $s_i \in s$ means that in time step $i$, the player has picked $i$ heroes, while the remaining part of the vector remains undetermined yet. Thus, the remaining dimensions will be filled with the default index $N_{classes}$. $N_{classes}$ is the total number of the hero pool.
        
    \item The action $a$ of the game: At the time-step $t$, the agent gives an action $a \in \mathcal{A} \subset [0, N_{classes} -1] \cap \mathbb{N}$, indicating that the player decides to pick the corresponding hero with the index range from $0$ to $N_{classes} -1$. 
    
    It is worth noting that not all the actions are legal, i.e., heroes picked in the past by the player himself or in the current rounds by both players are forbidden to be selected.
    
    \item The transition function $\mathcal{P}: \mathcal{S} \times \mathcal{A} \rightarrow \mathcal{S}$. $\mathcal{P}$ simply means that when taking an action $a_i$ at the state $s_i$, we get to a new state $s_{i+1}$.
    
    \item Single-round winning rate predictor $\phi$: Though we consider the full multi-round game as a whole, the winning rate of each round is only decided by the lineup of the heroes of an individual round. The winning rate of a single round here is estimated by the function $\phi$. It takes the lineups as input and outputs a winning rate $\{p, 1-p\}$ for each camp.
    
    \item The reward $r$: The two players get rewards $\{r, -r\}$ after the whole game ends. The reward is calculated by the following equation: $r = \sum_{i=0}^{D-1} \phi(lineup_i^1, lineup_i^2)$, where $lineup_i^1$ and $lineup_i^2$ are the lineups of $player1$ and $player2$ at round $i$.
    
    \item The goal of the game $\mathcal{G}$: To get a higher reward $r$ than the opponent player, i.e., get a positive reward $r$.

\end{itemize}

\subsection{Learning to Draft in MOBA Games with Neural Network and Tree Search}
In many \textit{two-player zero-sum games}, like Go, Chess, and Shogi, the family of Monte-Carlo tree search (MCTS) algorithms is seen as a powerful toolkit to get favorable actions with an acceptable computational cost. Recently, the variant of predictor upper confidence trees (PUCT) algorithm \cite{rosin2011multi} combined with deep neural networks, as known as AlphaZero, proposes a general framework for dealing with a series of chess games. However, to adapt to the draft problem in MOBA games, there are additional work to do. In this section, we will illustrate how we modify the mechanism of value back-propagation and adapt the framework to our multi-round draft games.

\begin{figure}[htbp]
    \centering
    \includegraphics[width=9cm]{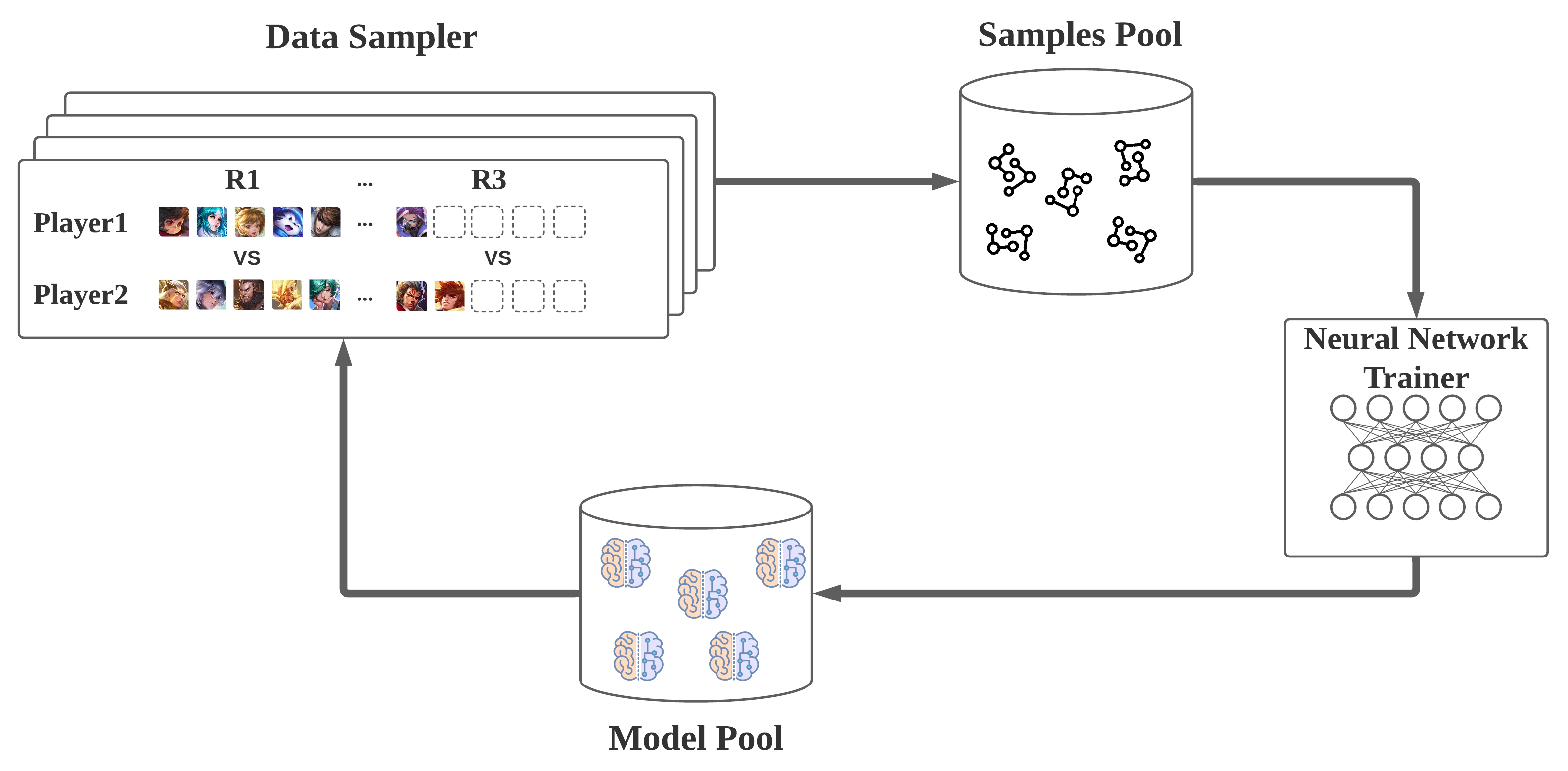}
    \caption{Overall Training Framework. The total framework consists of 4 modules: data sampler, sample pool, neural network trainer, model pool.}
    \label{fig:training_framework}
\end{figure}

\subsubsection{Overall Training Framework}
The overall training framework of JueWuDraft is shown in \fig{fig:training_framework}. To improve training efficiency, we make the framework distributed between a number of  CPU and GPU servers.

From the whole perspective, our framework consists of 4 modules: \textit{data sampler, samples pool, neural network trainer}, and \textit{model pool}. A \textit{data sampler} consists of two players and a gamecore, which generates samples by simulating matches between two players. The samples are sent to the \textit{samples pool} for training. The \textit{neural network trainer} performs supervised learning for the policy and value network with these samples and transmits the models to the \textit{model pool}, which synchronizes trained models to \textit{data samplers}. Take this as a loop, with the latest model, the data samplers continues to generate samples.


\subsubsection{Parallel MCTS with Deep Neural Networks}
In our framework, we utilize PUCT as the searching algorithm. Before introducing how to build a search tree, we first explain the meaning and the construction of the nodes in the tree.

In our algorithm, a node $n(s)$ is uniquely identified by a state $s$.
Usually, a node stores a set of statistics, $\{C(s), W(s), Q(s), P(s), VL(s)\}$. Here $C(s)$ is the visit count; $W(s)$ is the total value; $Q(s)$ is the mean value; $P(s)$ is the prior probability which is from the prediction of the policy network for its parent node; $VL(s)$ is the virtual loss for parallel tree searching \cite{chaslot2008parallel}. When a node is visited by a thread, the virtual loss increases. When the thread back-propagates the reward across this node, its virtual loss decreases. Thus, different threads are more likely to explore different nodes. The parent node $n(s)$ and the children node $n(s')$ are connected by edges $e(s,a)$, which means that we takes action $a$ at state $s$ and arrive at a new state $s'$.

\begin{figure*}[!t]
\centering
\includegraphics[width=\textwidth]{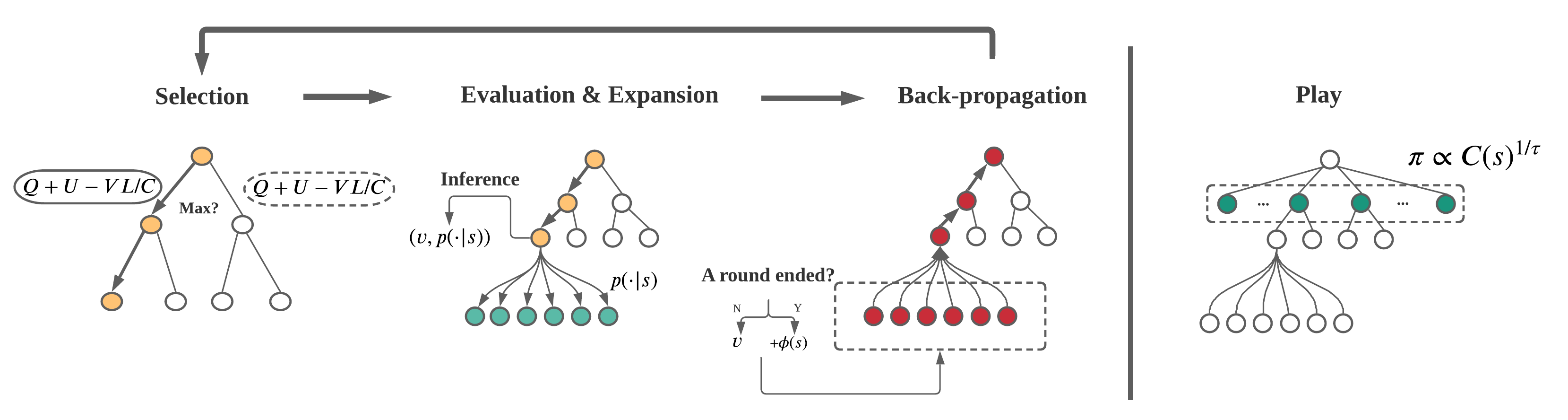}
\caption{MCTS for drafting process. Each simulation contains four steps: selection, evaluation, expansion and back-propagation, which are listed on the left. At the back-propagation step, the returned value will be added with a win-rate of this round when a round is end. After the search is complete, probabilities $\pi$ is returned (proportional to $C(s)$, the visit count of each node, and $\tau$ is a parameter controlling temperature), which is shown on the right.}
\label{fig:mcts}
\end{figure*}

To improve the speed of data sampling and fully utilize the computing resource, we perform searches on the tree in parallel.
Generally, our PUCT search tree is built iteratively in four steps: \textit{selection, evaluation, expansion} and \textit{back-propagation}. The process is illustrated in \fig{fig:mcts}.

\begin{itemize}
    \item \textit{\textbf{Selection}}. In this step, the algorithm begins at the root node that represents the current situation. Before meeting the leaf nodes of the search tree, the algorithm selects actions by the following equation and visit the deeper nodes:
    \begin{equation}
        a_t = \arg \max_{a}Q(s_{t,a}) + c_{puct}U(s_{t,a}) - VL(s_{t,a})/C(s_{t,a}), 
    \end{equation}
    \begin{equation}
         U(s_{t,a}) = P(s_{t,a})\sqrt{\sum_{a'}C(s_{t,a'})/C(s_{t,a})},
    \end{equation}
    where the mean value is calculated as $Q(s_t) = W(s_t)/C(s_t)$. $s_{t,a}$ means the new state $s_{t+1}$ after taking action $a$ at state $s_t$. Thus $Q(s_{t,a})$, $U(s_{t,a})$, $VL(s_{t,a}))$, $P(s_{t,a})$ and $C(s_{t,a})$ are equal to $Q(s_{t+1})$, $U(s_{t+1})$, $VL(s_{t+1})$, $P(s_{t+1})$, and $C(s_{t+1})$ respectively. For parallel searching, the virtual loss of the selected nodes should be added to avoid that different searching threads choose the same branch of the tree: $VL(s_{t, a}) \leftarrow VL(s_{t, a}) + c_{vl}$, where $c_{vl}$ is a hyperparameter for virtual loss. $c_{puct}$ is a balancing hyperparameter.
    
    \item \textit{\textbf{Evaluation}}. When the algorithm visits a leaf node $n(s_T)$, it stops step \textit{Selection} and turns to step \textit{Evaluation}. In \textit{Evaluation} step, the algorithm gets the oracle policy $\pi(\cdot|s_{T})$ and the value $f_\theta(s)$ by policy and value neural network through inference. 
    
    When a normal game (e.g., Go) ends, the winner is determined immediately. And the value is replaced with the win-lose result ($-1$ or $1$), which is designed as the ground-truth reward. However, we can only get complete lineups when the drafting process ends. Thus, we need to obtain the winning rate according to the complete lineup. As MOBA games are complex, computing the winning rate via taking an average of a bunch of games with the lineup is computationally infeasible.
    Therefore, we adopt the winning rate predictor $\phi(s)$ (we will introduce how it works later) to get a predicted winning rate and take it as the estimated value instead of the predicted value of the value network.
    
    Note that above we only replace the value when the last round ends, while there are several rounds during the multi-round game. The reason is that the estimation of our value network will have; a long-term consideration for both current rounds and following rounds. But the winning rate predictor only gets the value of the lineup for the current round.
    
    \item \textit{\textbf{Expansion}}. The algorithm expands the leaf node $n(s_T)$ as follows: Firstly, we mask all the illegal actions and renormalize the policy vector. Then according to the renormalized vector $\pi(\cdot|s_{T})$, for every legal action $a$, we create a new node $n(s_{T+1})$ and initialize all statistics to $0$, except $P(s_{t+1}) = \pi(a|s_{T})|_{s_{t+1} = s_{t,a}}$.
    
    \item \textit{\textbf{Back-propagation}}. After expansion, we update information for all nodes on the path from the node $n(s_T)$ to the root node. For every passed node $n(s)$, we update the statistics as $C(s) \leftarrow C(s) + 1$, $VL(s) \leftarrow VL(s) - c_{vl}$, and $Q(s) \leftarrow W(s)/C(s)$. For value update, the normal way is $W(s) \leftarrow W(s) + v(s_T)$. While in multi-round game, nodes in earlier rounds have influences on nodes in later rounds. Thus we will design a long-term value mechanism, which is introduced in subsection \textit{Long-term Value Propagation}.
\end{itemize}

\subsubsection{Policy and Value Network Training}

Providing the generalization ability to estimate value on unseen states, the deep neural network plays an important role in JueWuDraft, for both providing a stable reference of current value and dominant actions. Also, the neural network largely saves time when searching and building the tree as traditional MCTS algorithm samples to the end of the game, thus spending much time on rollouts. Thus, how to train the network is crucial.

\begin{figure}[htbp]
    \centering
    \includegraphics[width=0.5 \textwidth]{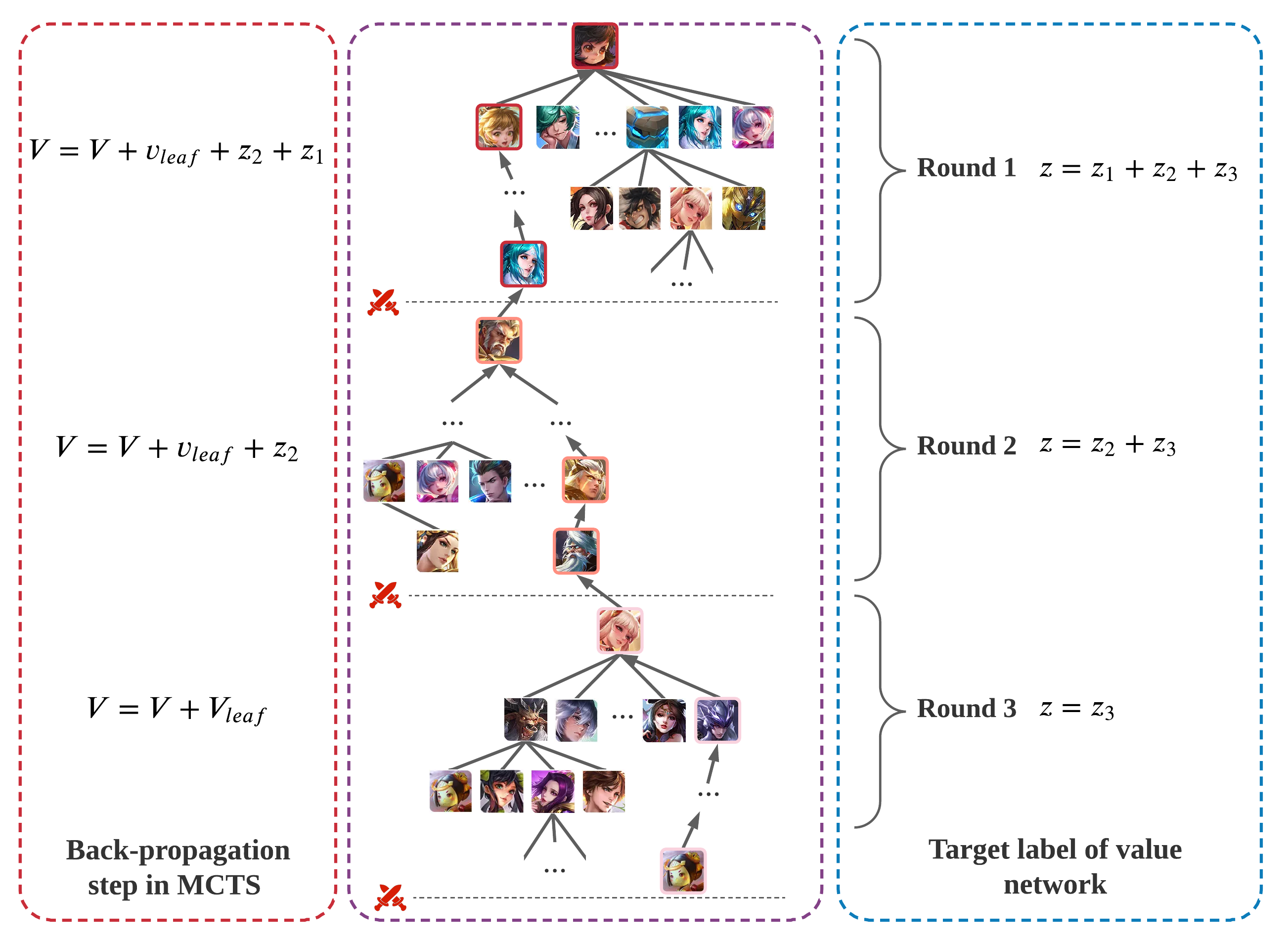}
    \caption{Value propagation}
    \label{fig:value_propagation}
\end{figure}

Our network outputs the value and oracle policy with a single unified architecture. Specifically, for a specific state $s$, the network $f_\theta$ parameterized by $\theta$ predicts the outcome $v_\theta(s)$ and policy $p_\theta(s)$ simultaneously. When training, we regard it as a supervised learning problem and adjust by gradient descent on a loss function $l$ that sums over mean-squared error and cross-entropy losses:
\begin{equation}
    l = (z-v_\theta(s))^2 - \pi ^ \top \log p_\theta(s) + c_p \| \theta \|^2,
\end{equation}
where $\pi$ is the returned probabilities by MCTS, shown on the right side of Fig. \ref{fig:mcts}. $c_p$ is the L2 penalty constant and $z$ is the target outcome in multi-round games.

 Firstly, since the winning rate predictor gives a winning rate $\phi(s) \in [0,1]$, which can be mapped to $[-1, 1]$ as reward signal:
\begin{equation}
    z_d = (\phi(s)-0.5)\times 2,\label{eq11}
\end{equation} 
where $z_d$ represents the transformed winning rate at the round $d$. While in multi-round games, past picked heroes cannot be chosen by the same player anymore, so the agents should make an advanced plan, and the target outcomes $z$ should be carefully designed, which will be discussed in the next subsection. 

\subsubsection{Long-term Value Propagation}
As discussed above, each drafting at earlier rounds will affect the outcome of later rounds in multi-round games. Also, it is intuitive to sum up all the outcomes in all related rounds for the value of the current step.
There are two cases of long-term value propagation:
\begin{itemize}
    \item For the \textit{Back-propagation} step in MCTS. Defining a node $n(s^d_t)$ on the path from the leaf node $n(s^D_T)$ to the root node $n(s^{D_0}_{T_0})$, $d$ is the round index and $t$ is the step index range from $T_0$ to $T$. The values of the node $n(s^d_t)$ and node $n(s^D_T)$ are $v(s^d_t)$ and $v(s^D_T)$, respectively. We update the value of the node $n(s^d_t)$ as:
\begin{equation}
    v^{new}(s^d_t) = v(s^d_t) + v(s^D_T) + \sum_{i \in [d, D)}{z_i} ,  
\end{equation}
    where $z_i$ is the transformed winning rate of round $i$, which is mentioned in Eq.~(\ref{eq11}). $v(s^D_T)$ only cares about values of round $D$ and further. Thus we add $\sum_{i \in [d, D)}{z_i}$ which reflects the influences of outcomes between round $d$ and round ${D-1}$.
    
    \item For the target label of the value network. For a step at round $d$, the target label is designed as 
    \begin{equation}\label{eqz}
        z = \sum_{i \geq d}z_i,
    \end{equation}
    which means the ground-truth value for one step at round $d$ only counts on current and later rounds, and it does not consider previous rounds. Each steps of the same player in the same round has the same target value. 
\end{itemize}

We show an example of a 3-round game in \fig{fig:value_propagation} for both the back-propagation step in MCTS and the target label for the value network. As is illustrated, nodes at earlier rounds affect more nodes than those later.

Note that the value back-propagation for $player1$ and $player2$ has the opposite sign since both players strive to maximize the predicted value of their team based on the current drafting situation.

\subsubsection{Network Structure and State Reconstruction}
In our multi-round draft game, the state $s$ is a vector with length $L \times D$. However, to train a policy and value network in a more efficient way, we need to reconstruct the state $s$.

\begin{figure}[htbp]
    \centering
    \includegraphics[width=0.48 \textwidth]{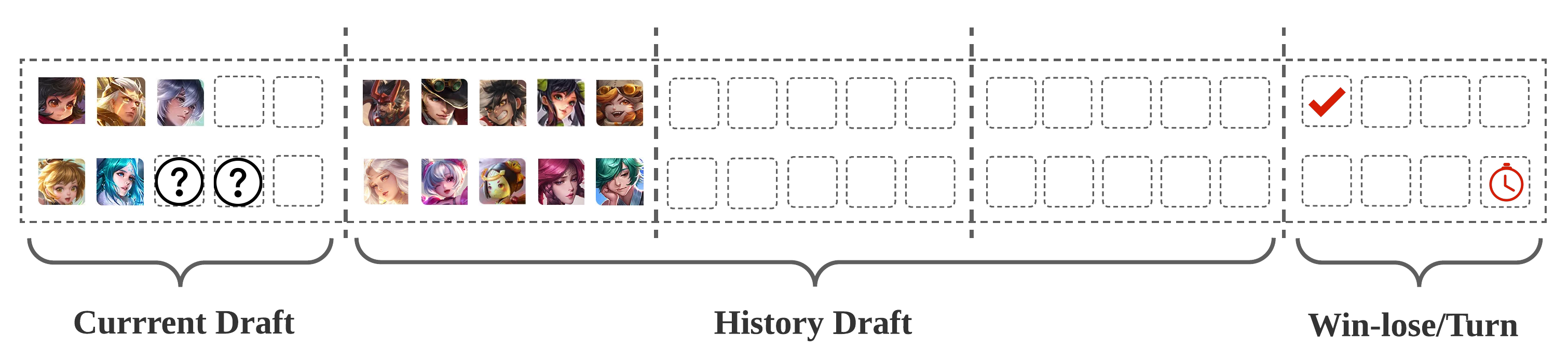}
    \caption{State vector configuration for policy and value network. It consists of three parts: heroes indexes of the current round; history draft information; the related game information (e.g., current round, whose turn.)}
    \label{fig:state_vector}
\end{figure}


\begin{figure*}[!t]
\centering
\subfigure[]{
\begin{minipage}[b]{0.42\linewidth}
\centering
\includegraphics[scale=0.6]{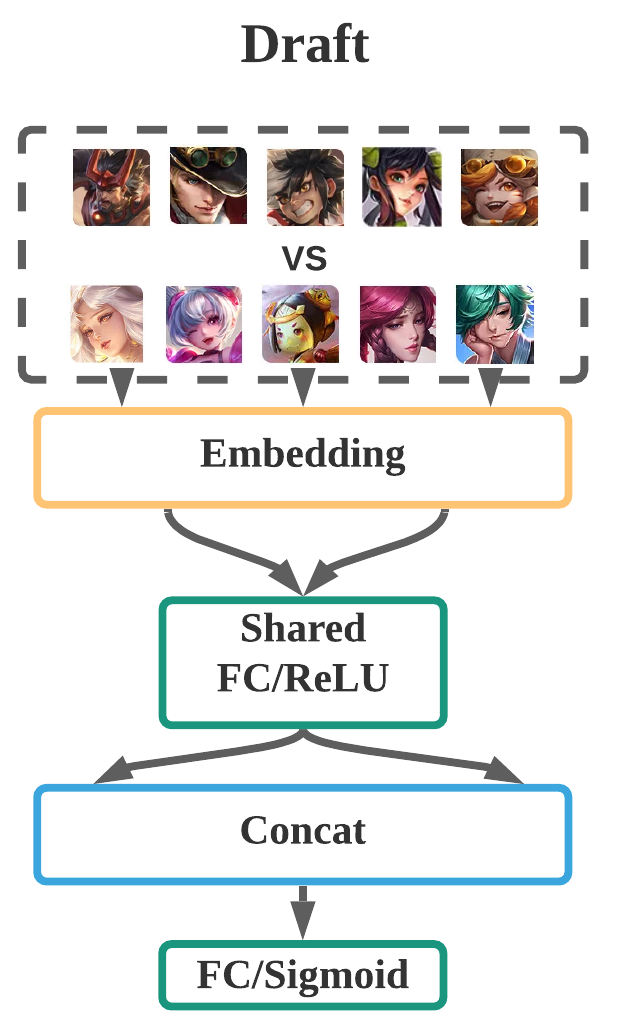}
\end{minipage}
}
\subfigure[]{
\begin{minipage}[b]{0.42\linewidth}
\centering
\includegraphics[scale=0.55]{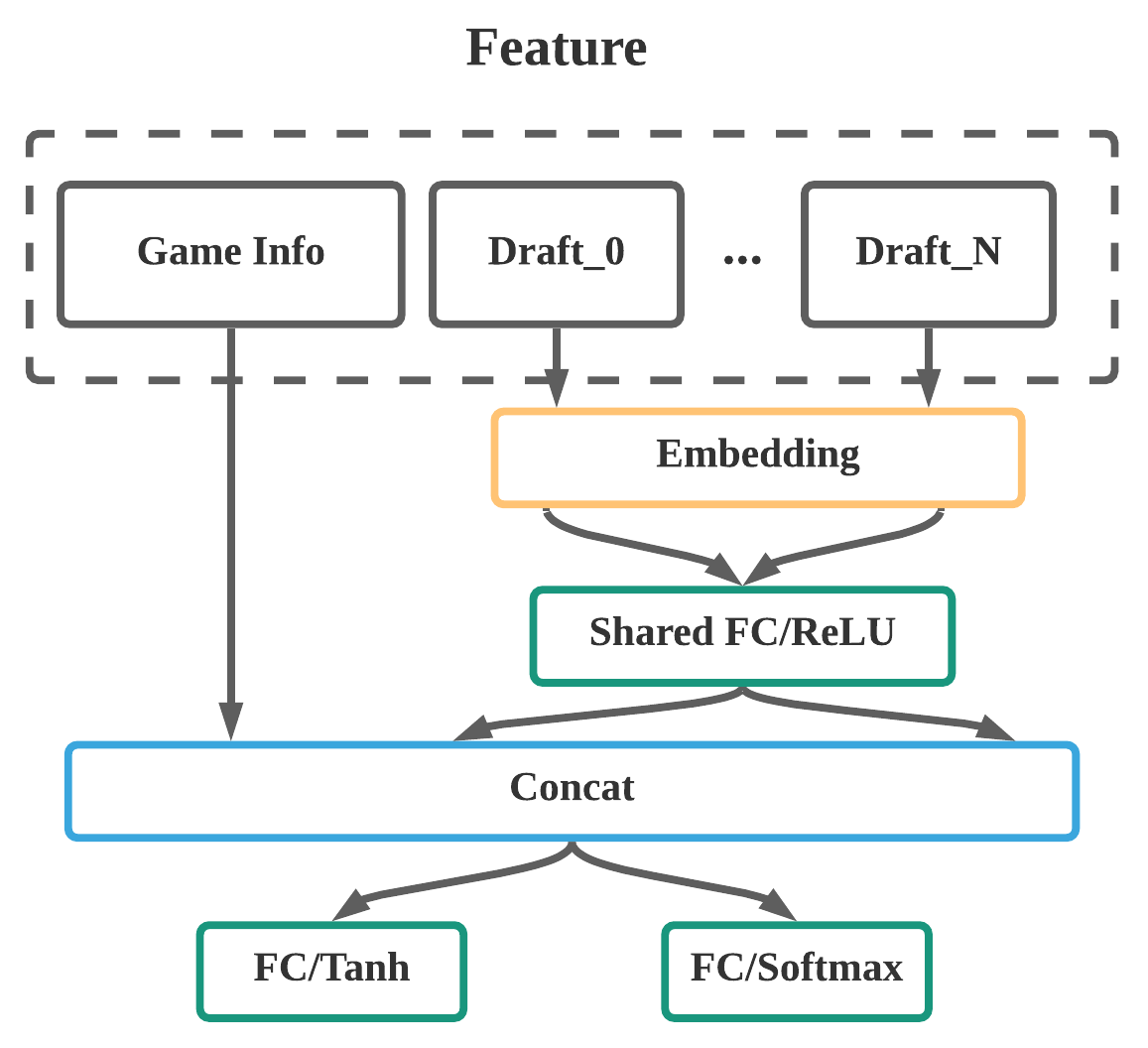}
\end{minipage}
}
\caption{Network Architecture for (a) the winning rate predictor and (b) the policy and value network. }
\label{fig:network}
\end{figure*}

As \fig{fig:state_vector} shows, the reconstructed vector mainly consists of three parts: the current draft part includes the draft in the current round, which mainly affects the winning rate. The middle of the vector is the history draft information. The last part is the related information (e.g., current round, whose turn to pick). Note that to further simplify the state, we here sort the hero index for each camp in current draft and history draft parts from small to large.

We use a simple 3-layer fully connected neural network as our policy and value network as shown in \fig{fig:network}(b). It takes in the state vector as shown in \fig{fig:state_vector} while outputs two heads: a fully-connected (FC) layer with a softmax function --- the policy head which outputs the probability of each action; an FC layer with a tanh function --- the value head which outputs the value of current state ranged from $-1$ to 1.

\subsection{Winning Rate Predictor}
We can only get complete lineups rather than win-lose information when the drafting process ends. Thus, we adopt the winning rate predictor to predict the winning rate of the lineup as our reward function. This winning rate predictor is trained with a match dataset which contains lineup and win-lose information. Each unique hero is distinguished with indices ranging from $[0, N_{classes}-1]$. The input feature is represented by the indices of all 10 picked heroes. 

The network architecture is shown in \fig{fig:network}(a), which is a simple 3-layer neural network with an output layer connected by a sigmoid activation function. Thus the output is the winning rate ranged from 0 to 1, which is always from the view of $camp1$.

%% file: IEEEtran/paper/experiment.tex
\section{Experiments}
\subsection{Datasets}
We train our winning rate predictor on the dataset containing a large number of matches in the game \textit{King of Honor}. In detail, we test JueWuDraft on two individual datasets, which are based on AI matches and human matches, respectively. Our simulation results are based on these two match datasets.

The AI Dataset: Note that we have developed an AI system to play full game of \textit{Honor of Kings}, which is discussed in \cite{ye2020towards}. Thus we collect matches via self-play of our trained AI during the training process when its performance becomes stable. The dataset contains 30 million matches, with each match containing hero lineup information and win-lose information, including 95 distinct heroes. 

The Human Dataset: We collect some matches of \textit{King of Honor} played by top 5\% human players  between August 1st, 2020 to August 10th, 2020. To be consistent with AI dataset, this match dataset is set to contain 30 million matches, with each match containing hero lineup information and win-lose information, including these 95 distinct heroes. 

\subsection{Winning Rate Predictor Performance}
To prepare the data for training the winning rate predictor, heroes of each match are distinguished with indices ranging in $[0, 94]$, while the match outcome is used as the label. The learning rate is set to $10^{-4}$ with Adam optimizer.

\begin{figure}[!t]
\centering
\subfigure[]{
\begin{minipage}[b]{\linewidth}
  \centering
  \includegraphics[]{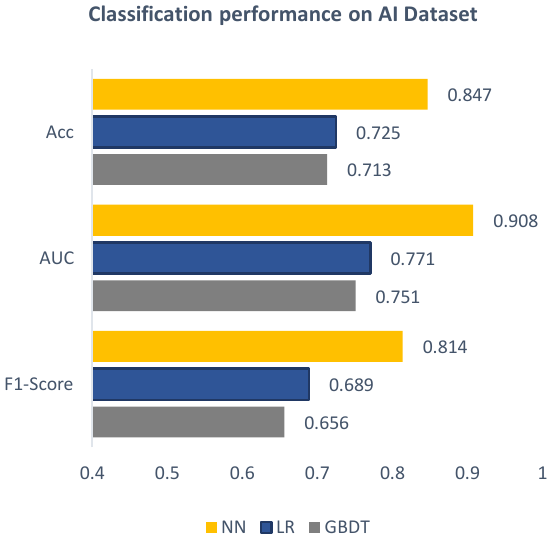}
  \label{fig:class_ai}
\end{minipage}
}

\subfigure[]{
\begin{minipage}[b]{\linewidth}
  \centering
  \includegraphics[]{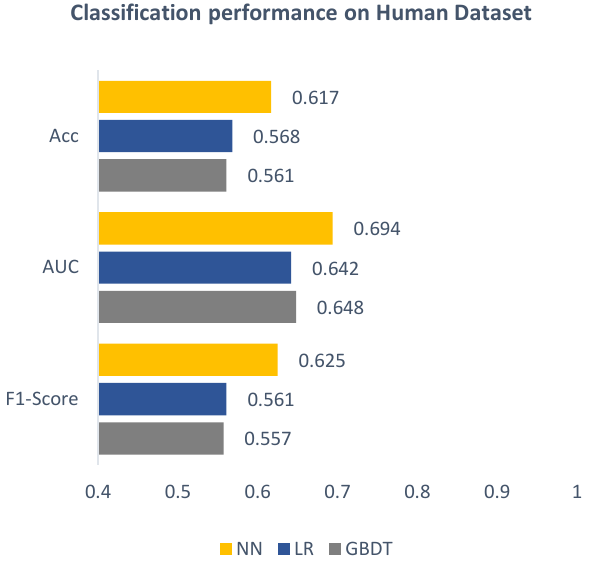}
  \label{fig:class_human}
\end{minipage}
}
\caption{Performance of classification methods for winning rate predictors for (a) the AI Dataset and (b) the Human Dataset.}
\end{figure}

We compare our winning rate predictor (neural network, NN) \cite{lecun1990handwritten,krizhevsky2012imagenet,simonyan2014very} with other two classification methods, including Gradient Boosted Decision Tree (GBDT) \cite{roe2005boosted,ye2009stochastic}, Logistic Regression (LR) \cite{wright1995logistic}. These methods are trained on two individual match datasets, respectively. And we report the accuracy, area under ROC curve (AUC), and the harmonic mean of the precision and recall (F1-Score) for each kind of methods in \fig{fig:class_ai} and \fig{fig:class_human}, averaged over 10-fold cross validation. 

As is shown in \fig{fig:class_ai} and \fig{fig:class_human}, NN performs the best on all metrics (accuracy, AUC and F1-Score) among these classification methods, which is 16.8\% better on the accuracy, 17.8\% better on AUC and 18.1\% better on F1-Score than the second best method (LR) for the AI Dataset, and 8.6\% better on the accuracy, 8.1\% better on AUC and 11.4\% better on F1-Score than the second best method (LR) for the Human Dataset, respectively. Thus we choose the designed neural network as our winning rate predictor. Note that all classification methods perform worse on the Human Dataset. The reason is that matches from the AI Dataset are played by similar policies. However, the Human Dataset contains matches from different human players, which makes it difficult to fit for all classification models. 

\subsection{Training Details.}
JueWuDraft is trained over a total number of 2400 CPU cores encapsulated in Dockers and a single Nvidia GPU (Tesla P40). The mini-batch size is 400; the learning rate is set to $10^{-4}$ with Adam optimizer; the $c_{vl}$ parameter is set to 3; the $c_{puct}$ parameter is 1; the temperature parameter $\tau$ is 1; and the L2 penalty parameter $c_p$ is $10^{-4}$;

For single-round game, about 0.4 million games of self-play (4 million samples) were generated during training, with 3200 iterations for each MCTS. The training process lasts about 2 hours until its loss converges. 

For 3-round game, about 0.4 million games of self-play (12 million samples) were generated during training, with 6400 iterations for each MCTS. The training time is about 24 hours. 

For 5-round game, about 0.3 million games of self-play (15 million samples) were generated during training, with 9600 iterations for each MCTS. The training time is about 32 hours. 

\subsection{Testing Details.}
We compare JueWuDraft with the other three strategies (DraftArtist, HWR, RD). To prove the effectiveness of long-term value mechanism for multi-round games, we also provide JueWuDraft strategy without long-term value mechanism (JueWuBase) as a comparison strategy. For every pair of strategies, we conduct 1000 simulation games: in each simulation, two players participate in the drafting process, with each player adopting one of the two compared strategies. At the end of each simulation, we collect the averaged winning rate predicted by $\phi(s) \in [0,1]$, with the 95\% confidence interval as a measure of strength of two strategies. Finally, the mean winning rates predicted by $\phi(s) \in [0,1]$, with the 95\% confidence interval over 1000 simulations of each pair of strategies are recorded and shown in table form. For each simulated draft, regardless of the strategy adopted, the first step of drafting heroes will be sampled from the probability distribution to cover more possible scenarios. And each strategy is set to pick heroes first for half of all 1000 simulations for fairness. All our test experiments are conducted on a PC with an Intel(R) Xeon(R) E5-2680 v4 $@$ 2.40GHz CPU (14 cores) and a single Nvidia GPU (Tesla M40).

The three comparison strategies are listed as below: 
\begin{itemize}
\item DraftArtist \cite{chen2018art}, which uses pure MCTS without our trained policy and value network. Note that this strategy will not consider situations of following rounds. 

\item Highest winning rate (HWR), selects the hero with the highest winning rate in the left hero pool, based on the statistical data. 

\item Random (RD), which selects the hero randomly in the remaining hero pool. 

\item JueWuDraft strategy without long-term value mechanism (JueWuBase), which uses MCTS with policy and value network, and without long-term value mechanism.
\end{itemize}

Note that we set the processing time for all MCTS-based methods to about 1 second for comparison, which is far lower than the time limit for gamecore (3 seconds) or limit for human players (30 seconds). And the processing time for MCTS-like method is mainly proportional to the number of iterations. According to processing time setting, the corresponding number of iterations will be $1.6\times 10^4$ for DraftArtist, $1.6\times 10^4$ for JueWuDraft (BO1), $1.3\times 10^4$ for JueWuDraft (BO3), and $1.0\times 10^4$ for JueWuDraft (BO5), respectively.

\begin{table}[!t]
\begin{minipage}[!t]{\columnwidth}
  \renewcommand{\arraystretch}{1.3}
  \caption{Mean predicted winning rate with the 95\% confidence interval of the row player in each pair of strategies for single-round game on the AI Dataset, where the row player plays against the column player. The strategies are sorted in ascending order by their strengths, from top to bottom, and from left to right. winning rates symmetric to diagonal always sum to one.}
  \label{table2}
  \centering
  \setlength{\tabcolsep}{0.6mm}{
  \begin{tabular}{ccccc}
  \toprule
  Row/Col.&RD&HWR&DraftArtist&JueWuDraft\\
  \midrule
  RD&-&-&-&-\\
  HWR&$0.982\pm0.004$&-&-&-\\
  DraftArtist&$0.983\pm0.006$&$0.591\pm0.005$&-&-\\
  JueWuDraft&$0.983\pm0.005$&$0.679\pm0.007$&$0.548\pm0.008$&-\\

  \bottomrule
  \end{tabular}}
  \end{minipage}
\\[12pt]
\begin{minipage}[!t]{\columnwidth}
  \renewcommand{\arraystretch}{1.3}
  \caption{Mean predicted winning rate with the 95\% confidence interval of each pair of strategies for 3-round game on the AI Dataset.}
  \label{table3}
  \centering
  \setlength{\tabcolsep}{0.8mm}{
  \begin{tabular}{ccccc}
  \toprule
  Row/Col.&RD&HWR&DraftArtist&JueWuBase\\
  \midrule
  RD&-&-&-&-\\
  HWR&$0.974\pm0.005$&-&-&-\\
  DraftArtist&$0.978\pm0.006$&$0.622\pm0.004$&-&-\\
  JueWuBase&$0.979\pm0.006$&$0.658\pm0.004$&$0.553\pm0.003$&-\\
  JueWuDraft&$0.981\pm0.005$&$0.677\pm0.003$&$0.580\pm0.004$&$0.516\pm0.003$\\
  \bottomrule
  \end{tabular}}
  \end{minipage}
  \vspace{-0.1cm}
\\[12pt]
\begin{minipage}[!t]{\columnwidth}
  \renewcommand{\arraystretch}{1.3}
  \caption{Mean predicted winning rate with the 95\% confidence interval of each pair of strategies for 5-round game on the AI Dataset.}
  \label{table4}
  \centering
  \setlength{\tabcolsep}{0.8mm}{
  \begin{tabular}{ccccc}
  \toprule
  Row/Col.&RD&HWR&DraftArtist&JueWuBase\\
  \midrule
  RD&-&-&-&-\\
  HWR&$0.938\pm0.007$&-&-&-\\
  DraftArtist&$0.955\pm0.009$&$0.594\pm0.011$&-&-\\
  JueWuBase&$0.966\pm0.006$&$0.629\pm0.005$&$0.570\pm0.005$&-\\
  JueWuDraft&$0.970\pm0.006$&$0.644\pm0.005$&$0.601\pm0.006$&$0.528\pm0.005$\\
  \bottomrule
  \end{tabular}}
  \end{minipage}
  \vspace{-0.1cm}
\end{table}

\begin{table}[!t]
\begin{minipage}[!t]{\columnwidth}
  \renewcommand{\arraystretch}{1.3}
  \caption{Mean predicted winning rate with the 95\% confidence interval of each pair of strategies for single-round game on the Human Dataset, where the row player plays against the column player.  The strategies are sorted in ascending order by their strengths, from top to bottom, and from left to right. Winning rates symmetric to diagonal always sum to one.}
  \label{table21}
  \centering
  \setlength{\tabcolsep}{0.6mm}{
  \begin{tabular}{ccccc}
  \toprule
  Row/Col.&RD&HWR&DraftArtist&JueWuDraft\\
  \midrule
  RD&-&-&-&-\\
  HWR&$0.646\pm0.006$&-&-&-\\
  DraftArtist&$0.699\pm0.005$&$0.537\pm0.007$&-&-\\
  JueWuDraft&$0.756\pm0.005$&$0.600\pm0.002$&$0.525\pm0.003$&-\\
  \bottomrule
  \end{tabular}}
  \end{minipage}
\\[12pt]
\begin{minipage}[!t]{\columnwidth}
  \renewcommand{\arraystretch}{1.3}
  \caption{Mean predicted winning rate with the 95\% confidence interval of each pair of strategies for 3-round game on the Human Dataset.}
  \label{table31}
  \centering
  \setlength{\tabcolsep}{0.8mm}{
  \begin{tabular}{ccccc}
  \toprule
  Row/Col.&RD&HWR&DraftArtist&JueWuBase\\
  \midrule
  RD&-&-&-&-\\
  HWR&$0.595\pm0.005$&-&-&-\\
  DraftArtist&$0.693\pm0.005$&$0.571\pm0.006$&-&-\\
  JueWuBase&$0.719\pm0.006$&$0.618\pm0.005$&$0.533\pm0.005$&-\\
  JueWuDraft&$0.739\pm0.006$&$0.642\pm0.006$&$0.558\pm0.005$&$0.520\pm0.006$\\
  \bottomrule
  \end{tabular}}
  \end{minipage}
  \vspace{-0.1cm}
\\[12pt]
\begin{minipage}[!t]{\columnwidth}
  \renewcommand{\arraystretch}{1.3}
  \caption{Mean predicted winning rate with the 95\% confidence interval of each pair of strategies for 5-round game on the Human Dataset.}
  \label{table41}
  \centering
  \setlength{\tabcolsep}{0.8mm}{
  \begin{tabular}{ccccc}
  \toprule
  Row/Col.&RD&HWR&DraftArtist&JueWuBase\\
  \midrule
  RD&-&-&-&-\\
  HWR&$0.587\pm0.010$&-&-&-\\
  DraftArtist&$0.675\pm0.013$&$0.589\pm0.007$&-&-\\
  JueWuBase&$0.698\pm0.008$&$0.609\pm0.007$&$0.523\pm0.006$&-\\
  JueWuDraft&$0.712\pm0.007$&$0.626\pm0.006$&$0.547\pm0.005$&$0.525\pm0.006$\\
  \bottomrule
  \end{tabular}}
  \end{minipage}
  \vspace{-0.1cm}
\end{table}
\begin{table*}[!htbp]
\centering
\caption{Average process time of picking a hero for all methods (in seconds).}
\begin{tabular}{cccccc}
\toprule
RD&HWR&DraftArtist&JueWuDraft\\
\midrule
$2\times 10^{-5}$ & $2\times 10^{-5}$&1.0&1.0\\
\bottomrule
\end{tabular}
\label{table5}
\end{table*}

\subsection{Simulation Results}
\subsubsection{Effectiveness} We compared JueWuDraft with other three strategies (DraftArtist, HWR, RD) and a comparison strategy (JueWuBase) trained on both two match datasets (the AI Dataset and the Human Dataset). And results of both single-round and multi-round games (BO3, BO5) are provided. Each cell in the table means the mean predicted winning rate $\phi(s) \in [0,1]$ with the 95\% confidence interval which is played by each pair of strategies (the corresponding strategy from the first column and the corresponding strategy from the first row). We judge the strength of one strategy by analyzing the mean predicted winning rates against other strategies. For example, one strategy can defeat another strategy with more than 50\% averaged winning rate, we consider it is a better strategy. The predicted winning rates shown in the table are from the view of strategies in the first column. In total, we present the results in six tables (namely, Table~ \ref{table2}-\ref{table41}), each of which represents one specific experimental setting.

\minisection{Single-round game (BO1)} JueWuDraft beats the second place (DraftArtist) with 54.8\% on the AI Dataset as shown in Table \ref{table2}, and 52.5\% better than the second place on the Human Dataset shown in Table \ref{table21}, respectively. It shows that JueWuDraft performs best among all strategies for single-round game.

\minisection{3-round game (BO3)} Experiments show that JueWuDraft outperforms other strategies for 3-round game. It has a mean predicted winning rate of 58.0\%  against the second place on the AI Dataset shown in Table \ref{table3}, and 55.8\% better than the second place on the Human Dataset shown in Table \ref{table31}, respectively. And with long-term value mechanism, the mean predicted winning rates against DraftArtist increased by 4.9\% on the AI Dataset and 4.7\% on the Human Dataset, respectively.

\minisection{5-round game (BO5)} JueWuDraft has a mean predicted winning rate of 60.1\% against the second place on the AI Dataset shown in Table \ref{table4}, and 54.7\% better than the second place on the Human Dataset shown in Table \ref{table41}, respectively. It demonstrates that JueWuDraft achieves the best performance for 5-round game. And JueWuDraft (with the long-term value mechanism) beats over JueWuBase (without long-term value mechanism) over 52.8\% on AI Dataset, and 52.5\% on Human Dataset, respectively.

Comparing performance between single-round and multi-round games (BO3, BO5), it can be seen that JueWuDraft performs better against the baseline method (DraftArtist) on multi-round games. The main reason is that the multi-round games are more complex and need more consideration for both current states and the following rounds. While JueWuDraft can handle further consideration better than the other strategies. 

Noted that JueWuDraft achieves higher predicted winning rates for the AI Dataset compared to these for the Human Dataset. And the differences of average predicted winning rates between each pair of strategies are smaller for the Human Dataset.
One possible reason is that matches from the AI Dataset are collected via self-play of similar policies, enlarging the gaps in winning rates for heroes. While matches from the Human Dataset are collected from various kinds of human players at a similar level, which will weaken the effects of lineups to game results. According to our statistics, the mean winning rates of heroes range from 41.4\% to 55.4\% for the Human Dataset, and the interval is [26.4\%, 81.9\%] for the AI Dataset.

\subsubsection{Efficiency}

To show the efficiency of different strategies, we provide the average processing time per step of the drafting process for every strategy in Table \ref{table5}. The average time of picking a hero by JueWuDraft is about 1 second. It is acceptable to apply our drafting method to the game in real time. The time limit for our gamecore is 3 seconds, and it is over 30 seconds for human players. Therefore, our drafting system is suitable for being applied in real game with a large scale. 

\begin{figure}[htbp]
    \centering
    \includegraphics[width=0.45\textwidth]{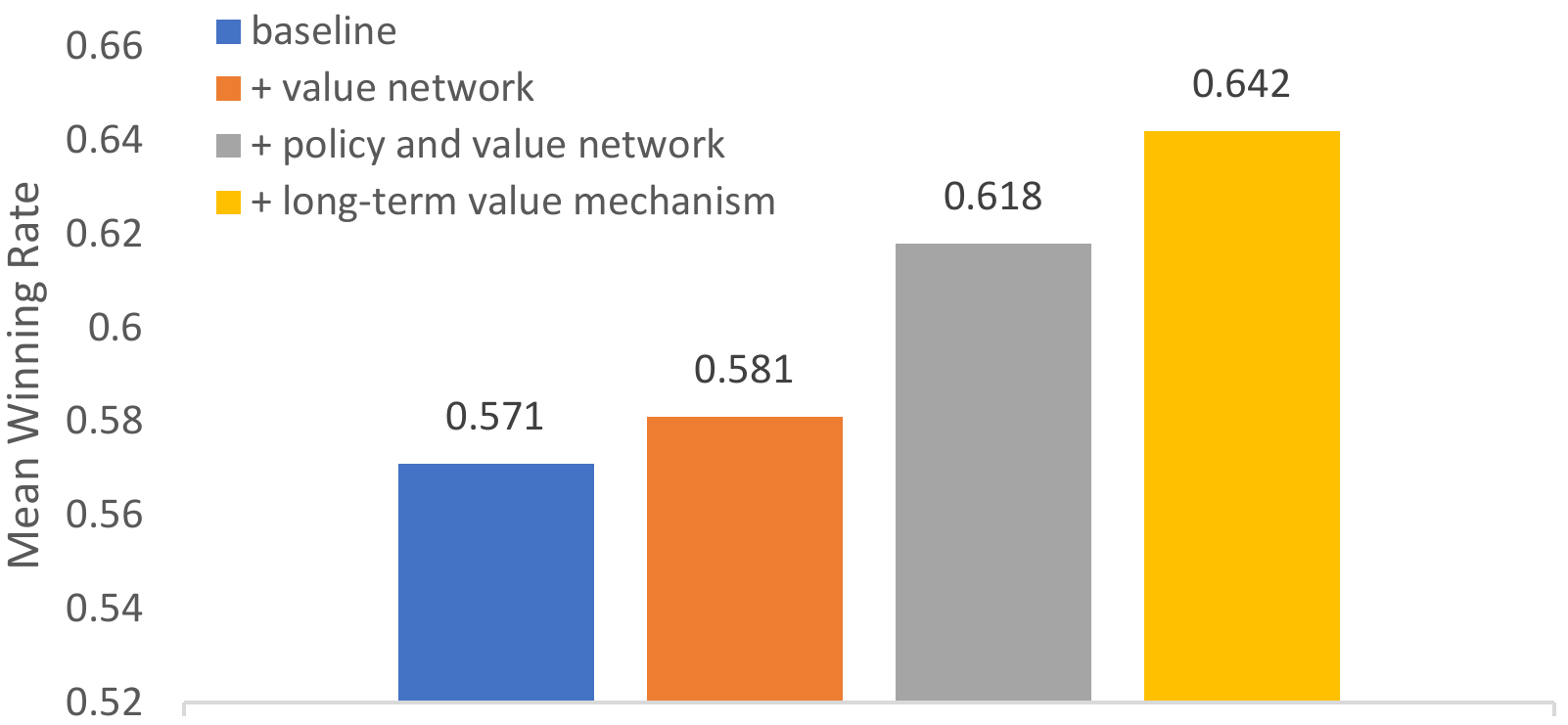}
    \caption{Ablation study for key components.These experiments use a specific setup: 3-round game on Human Dataset, taking HWR as a comparison method.}
    \label{fig:ablation}
\end{figure}

\begin{figure*}[!t]
    \centering
    \includegraphics[width=\textwidth]{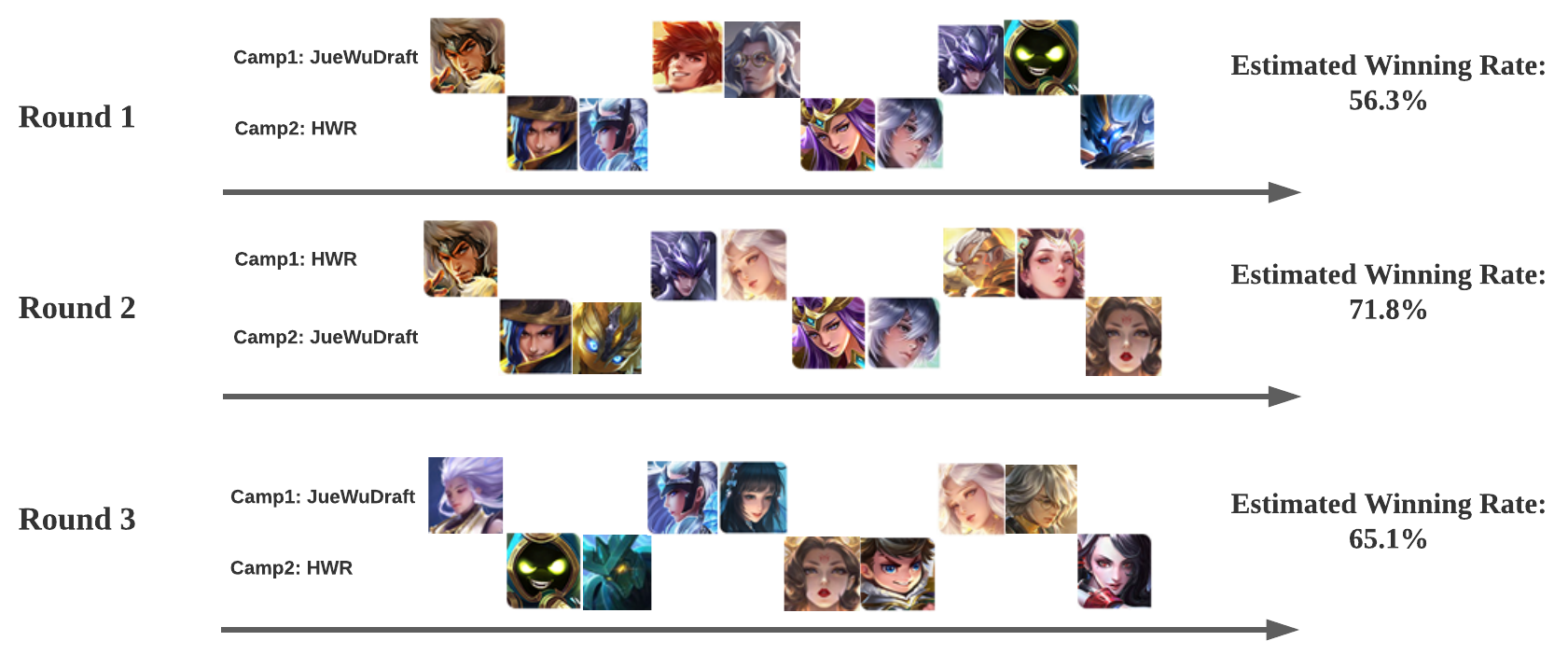}
    \caption{One specific simulation case of JueWuDraft against with HWR strategy, where JueWuDraft takes the first move.}
    \label{fig:case}
\end{figure*}

\subsubsection{Ablation Study}
To further analyze the key components in JueWuDraft, we run an internal ablation, which is shown in \fig{fig:ablation}. We test JueWuDraft by adding key components gradually with a specific setting: 1000 simulations for 3-round game on Human Dataset, taking HWR as its opponent strategy. The baseline in \fig{fig:ablation} is the MCTS strategy, and three other strategies are MCTS with the value network (a provisional version of JueWuDraft used in \cite{ye2020towards}), MCTS with policy and value network (JueWuBase), MCTS with policy and value network and long-term value mechanism (the full version of JueWuDraft). It shows that using value network can increase the mean predicted winning rate slightly compared with the baseline strategy. Then, adding policy and value network obtains better results significantly. Further, we observe that the performance improves after adding long-term value mechanism. It demonstrates that adding policy network, value network and long-term value mechanism, all of them are helpful for our drafting process.

Here we show one specific simulation case of our final method, against with HWR strategy, which is a greedy strategy and tends to choose the hero with the highest predicted winning rates. The predicted winning rates for JueWuDraft in three rounds are 56.3\%, 71.8\%, 65.1\%, respectively. The predicted winning rates are higher in the second and third rounds rather than in the first round. Specifically, JueWuDraft chooses hero \textit{Chengjisihan} rather than \textit{Liubei} (the hero with highest mean predicted winning rate) at the first hand, although choosing \textit{Liubei} firstly may get a higher predicted winning rate in the first round. 
It indicates that JueWuDraft is a less greedy strategy and considers longer for later rounds.  

In conclusion, JueWuDraft outperforms other strategies in all experimental settings, which demonstrates both the effectiveness and the efficiency of our algorithm.

%% file: IEEEtran/paper/conclusion.tex
\section{Conclusion and Future Work}
In this paper, we propose a novel drafting algorithm based on Monte-Carlo tree search and neural networks. Furthermore, we formulate the multi-round drafting process as a combinatorial game, and design a long-term value mechanism to adapt our method to the best-of-$N$ drafting problem. Extensive experimental results show the effectiveness and efficiency of our proposed algorithm, as compared with other drafting methods.

The limitations of the work so far can be summarized in three aspects. 

Firstly, besides the drafting process, the process of banning heroes is also an important pre-match phase, which is also known as Ban/Pick process. Yet in this paper we have not considered about banning heroes yet in this algorithm. 

Secondly, the training datasets are collected from RL training matches or human matches, while the whole AI system is built to fight against human players, which forms a gap in winning rates between our trained agents and human players. However, our designed winning rate predictor does not explicitly consider this gap. Also, different human teams have their own playing style, which cannot be modeled by a single win-rate predictor that trained by match results played by different top human players. How to adapt to different MOBA opponent for our win-rate predictor, it's an unsolved problem in this paper. 

Finally, the rule of multi-round games is that a player needs only taking over half rounds of all rounds to become a winner. We have tried to design specific value objective function to maximize the probability of winning corresponding rounds (e.g., 2 of 3 rounds). However, we find that value mechanism is hard to convergence for our network. Thus, we choose a easy and simple way (maximize the mean winning rates of all rounds) as our value mechanism, which is a imperfect approximation of what we really want. In general, the value credit assignment for multi-round drafting problem is still a problem worth to be explored.

In the future, we will work on these three aspects to further improve JueWuDraft.

\section{Broader Impact}
\minisection{To the research community} MOBA game is a challenging field for all AI researchers. Building an AI system to play full MOBA 5v5 games with over 100 heroes in a best-of-$N$ competition is still an unconquered but significant task for AI community. Hero drafting is a necessary pre-match procedure for MOBA game. Thus, developing an automatic hero drafting algorithm is an important part of a complete AI system for MOBA game. Speaking of this paper, it proposes a hero drafting method leveraging Monte-Carlo tree search and neural network. As far as we know, it is the first time that best-of-$N$ drafting problem is considered, where we formulate it as a multi-round combinatorial game, and design a long-term value mechanism to adapt to it. The experiment results demonstrate the superiority of our method. We herewith expect this work to provide inspiration to other complex real-world problems, e.g., real-time multi-step decisions of robotics.

\minisection{To the game industry} \textit{Honor of Kings} is the most popular MOBA game all over the world, also the most downloaded App worldwide \footnote{https://en.wikipedia.org/wiki/Honor\_of\_Kings}. As an important part of a complete AI system for MOBA game, our automatic hero drafting method has found several real-world applications in the game and is changing the way of playing MOBA games. These applications can be concluded as below:
\begin{enumerate}
    \item PVE (player vs environment) game mode. We have developed a whole AI system which can be used for \textit{Honor of Kings} in PVE game mode \cite{ye2020towards}, where JueWuDraft can be used for picking heroes for AI camp. 
    \item Hero recommendation for human players. At the pre-match phase of games, JueWuDraft can provide a prior hero list and an inferior hero list to help human players to pick heroes that can cooperate well with teammates and counter with opponents. 
\end{enumerate}